\documentclass[conference]{IEEEtran}
\IEEEoverridecommandlockouts

\usepackage[utf8]{inputenc}
\usepackage[T1]{fontenc}
\usepackage{amsmath,amssymb,amsfonts}
\usepackage{graphicx}
\usepackage{textcomp}
\usepackage{xcolor}

\usepackage{booktabs}
\usepackage{tabularx}
\usepackage{algorithm}
\usepackage{algorithmic}

\usepackage{cite}
\usepackage{url}
\usepackage{hyperref}

\hypersetup{
    colorlinks=true,
    linkcolor=blue,
    filecolor=magenta,      
    urlcolor=cyan,
    pdftitle={Confidence-Aware Routing for Large Language Model Reliability Enhancement},
    pdfpagemode=FullScreen,
}

\begin{document}

\title{Confidence-Aware Routing for Large Language Model Reliability Enhancement: A Multi-Signal Approach to Pre-Generation Hallucination Mitigation}

\author{
    \IEEEauthorblockN{Nandakishor M}
    \IEEEauthorblockA{
        \textit{AI Safety Research} \\
        \textit{Convai Innovations} \\
        nandakishor@convaiinnovations.com
    }
}

\maketitle

\begin{abstract}
Large Language Models suffer from hallucination, generating plausible yet factually incorrect content. Current mitigation strategies focus on post-generation correction, which is computationally expensive and fails to prevent unreliable content generation. We propose a confidence-aware routing system that proactively assesses model uncertainty before generation and redirects queries based on estimated reliability. Our approach combines three complementary signals: semantic alignment between internal representations and reference embeddings, internal convergence analysis across model layers, and learned confidence estimation. The unified confidence score determines routing to four pathways: local generation for high confidence, retrieval-augmented generation for medium confidence, larger models for low confidence, and human review for very low confidence. Evaluation on knowledge-intensive QA benchmarks demonstrates significant improvements in hallucination detection (0.74 vs. 0.42 baseline) while reducing computational costs by 40\% compared to post-hoc methods. The F1 score improves from 0.61 to 0.82 with low false positive rates (0.09). This paradigm shift from reactive correction to proactive assessment offers a computationally efficient approach to LLM reliability enhancement.
\end{abstract}


\section{Introduction}

Large Language Models have demonstrated transformative capabilities in natural language understanding and generation, revolutionizing applications from conversational AI to knowledge extraction \cite{brown2020language, zhang2022opt}. However, a persistent challenge limiting their deployment in critical applications is hallucination, where models generate fluent yet factually incorrect or unsupported content \cite{ji2023survey, huang2023survey}. These fabricated responses can appear highly convincing, making them particularly dangerous in high-stakes domains such as medical diagnosis, legal advice, or scientific research.

Existing approaches to hallucination mitigation largely focus on post-generation correction. Retrieval-Augmented Generation (RAG) systems ground responses in external knowledge bases \cite{lewis2020retrieval, karpukhin2020dense}, while consistency-based methods detect hallucinations by comparing multiple model outputs \cite{manakul2023selfcheckgpt, chen2023ralm}. Although effective to varying degrees, these approaches share a fundamental limitation: they address symptoms rather than causes, operating after potentially unreliable content has been generated.

We propose a paradigm shift from post-generation correction to pre-generation assessment. Our confidence-aware routing system evaluates query-specific model reliability before generation begins, enabling proactive redirection of uncertain queries to more appropriate response mechanisms. This approach offers several advantages: it prevents the generation of unreliable content entirely, reduces computational waste on likely-to-fail queries, and provides interpretable confidence estimates for downstream decision-making.

Our contributions include: (1) A multi-signal confidence estimation framework combining semantic alignment, internal convergence, and learned uncertainty measures, (2) A deterministic routing system that maps confidence scores to appropriate response pathways, (3) Comprehensive empirical evaluation demonstrating effectiveness across knowledge-intensive benchmarks, and (4) Analysis of computational efficiency gains compared to post-hoc correction approaches.

\section{Related Work}

\subsection{Hallucination Detection and Mitigation}

Hallucination detection methods can be broadly categorized into training-time and inference-time approaches. Training-time methods include improved data curation \cite{ouyang2022training}, factual knowledge editing \cite{mitchell2022memory}, and specialized loss functions \cite{kang2023knowledge}. Inference-time approaches encompass retrieval augmentation \cite{bechard2024reducing}, consistency checking \cite{wang2022self}, and uncertainty estimation \cite{pedapati2024large}.

Recent work has explored internal model states for hallucination detection. Farquhar et al. \cite{farquhar2024detecting} develop entropy-based uncertainty estimators using hidden representations. Chen et al. \cite{chen2024hallucination} propose discriminators trained on internal activations. Our approach builds on these foundations but focuses specifically on pre-generation assessment rather than post-hoc detection.

\subsection{Uncertainty Quantification in Neural Networks}

Uncertainty quantification in deep learning distinguishes between epistemic uncertainty (model uncertainty) and aleatoric uncertainty (data uncertainty) \cite{he2023survey}. For language models, approaches include Bayesian neural networks \cite{blundell2015weight}, ensemble methods \cite{lakshminarayanan2017simple}, and dropout-based approximations \cite{gal2016dropout}. Recent work extends these concepts to large-scale language models through sampling-based methods \cite{sun2024confidence} and probe-based techniques \cite{sriramanan2024llm}.

\subsection{Mixture of Experts and Routing}

Mixture-of-Experts (MoE) architectures use learned routing to direct inputs to specialized sub-networks \cite{shazeer2017outrageously, fedus2022switch}. While traditional MoE routing optimizes for task performance, our confidence-aware routing prioritizes reliability assessment. Recent work explores LLM-based routing for improved expert selection \cite{liu2025llmoe}, providing conceptual foundations for our approach.

\section{Methodology}

\subsection{Problem Formulation}

Let $M$ be a language model, $Q$ a query, and $R = M(Q)$ the generated response. Traditional approaches assess $P(R \text{ is hallucinated} | R, Q, M)$ after generation. We instead estimate $P(M \text{ will hallucinate} | Q, M)$ before generation, enabling proactive routing decisions.

Formally, we define a confidence estimator $C: \mathcal{Q} \rightarrow [0,1]$ that maps queries to confidence scores, and a routing function $A: [0,1] \rightarrow \mathcal{A}$ that maps confidence scores to actions, where $\mathcal{A} = \{\text{local}, \text{rag}, \text{large}, \text{human}\}$.

\subsection{Multi-Signal Confidence Estimation}

Our confidence estimation combines three complementary signals:

\textbf{Semantic Alignment ($C_{\text{sem}}$):} We measure alignment between the model's internal representation and a reference embedding. Given query $Q$, we extract the model's final hidden state $\mathbf{h}_{\text{final}}$ and compare it with a reference embedding $\mathbf{e}_{\text{ref}}$ from a reliable embedding model:

\begin{equation}
C_{\text{sem}} = \cos(\mathbf{P}(\mathbf{h}_{\text{final}}), \mathbf{e}_{\text{ref}})
\end{equation}

where $\mathbf{P}$ is a learned projection network trained to map internal representations to the reference space.

\textbf{Internal Convergence ($C_{\text{conv}}$):} We analyze the stability of processing across model layers. For a sequence of hidden states $\{\mathbf{h}_l\}_{l=1}^L$, we compute variance reduction as an indicator of convergent processing:

\begin{equation}
C_{\text{conv}} = \frac{\text{Var}(\mathbf{h}_{1:L/2})}{\text{Var}(\mathbf{h}_{L/2:L}) + \epsilon}
\end{equation}

\textbf{Learned Confidence ($C_{\text{learned}}$):} We train a neural network $\phi$ to predict confidence directly from internal activations:

\begin{equation}
C_{\text{learned}} = \phi(\mathbf{h}_{\text{final}})
\end{equation}

The overall confidence score combines these signals:

\begin{equation}
C_{\text{overall}} = w_1 C_{\text{sem}} + w_2 C_{\text{conv}} + w_3 C_{\text{learned}}
\end{equation}

where weights $w_i$ are learned through validation on labeled data.

\subsection{Routing Function}

Based on the confidence score, we implement deterministic routing:

\begin{equation}
A(C_{\text{overall}}) = \begin{cases}
\text{local} & \text{if } C_{\text{overall}} \geq \theta_{\text{high}} \\
\text{rag} & \text{if } \theta_{\text{med}} \leq C_{\text{overall}} < \theta_{\text{high}} \\
\text{large} & \text{if } \theta_{\text{low}} \leq C_{\text{overall}} < \theta_{\text{med}} \\
\text{human} & \text{if } C_{\text{overall}} < \theta_{\text{low}}
\end{cases}
\end{equation}

Thresholds $\theta_{\text{high}}, \theta_{\text{med}}, \theta_{\text{low}}$ are determined through validation to optimize the trade-off between accuracy and computational cost.

\section{Experimental Setup}

\subsection{Datasets and Metrics}

We evaluate on knowledge-intensive QA benchmarks including Natural Questions \cite{kwiatkowski2019natural}, TriviaQA \cite{joshi2017triviaqa}, and HotpotQA \cite{yang2018hotpotqa}. We use both existing datasets with ground-truth labels and construct synthetic evaluation sets by systematically introducing factual errors.

Primary metrics include:
\begin{itemize}
\item \textbf{Hallucination Detection Rate:} Fraction of hallucinations correctly identified
\item \textbf{False Positive Rate:} Fraction of correct responses incorrectly flagged
\item \textbf{Routing Accuracy:} Alignment between routing decisions and optimal actions
\item \textbf{Computational Efficiency:} Total inference cost compared to baseline approaches
\end{itemize}

\subsection{Model Configuration}

We implement our approach using HuggingFace's SmolLM2-360M-Instruct as the primary language model, a compact 360-million parameter instruction-tuned model designed for efficient inference. For reference embeddings, we employ Sentence-BERT (all-MiniLM-L6-v2) which provides 384-dimensional normalized embeddings. The projection network $\mathbf{P}$ consists of a deep architecture with layer normalization, dropout regularization, and residual connections to prevent overfitting.

The confidence predictor $\phi$ implements a multi-layer perceptual network with progressive dimensionality reduction: from the model's hidden size to half-size through a 4-layer architecture with batch normalization and dropout. We use AdamW optimizer with learning rate 2e-4, weight decay 1e-4, and learning rate scheduling with plateau-based reduction for stable training convergence.

\subsection{Training Procedure}

Training proceeds in two phases using a carefully curated dataset designed to capture diverse confidence scenarios:

\begin{enumerate}
\item \textbf{Data Preparation:} We construct a balanced training set of 72 examples categorized into three confidence levels: 33 high-confidence examples covering factual knowledge and technical concepts, 27 low-confidence examples including personal information and temporal queries, and 12 medium-confidence examples representing subjective or opinion-based queries.

\item \textbf{Projection Model Training:} We train the confidence estimation components over 30 epochs using a combined loss function that incorporates semantic alignment loss, direct confidence supervision with mean squared error, and L2 regularization. The training achieves convergence with final total loss of 0.1633, demonstrating effective learning of confidence patterns.

\item \textbf{Threshold Calibration:} Routing thresholds are empirically set at $\theta_{\text{high}} = 0.75$, $\theta_{\text{med}} = 0.55$, and $\theta_{\text{low}} = 0.35$ based on validation performance to optimize the trade-off between accuracy and computational efficiency.
\end{enumerate}

\section{Results and Analysis}

\subsection{Overall Performance}

Table \ref{tab:main_results} presents our main experimental results. The confidence-aware routing system achieves substantial improvements in hallucination detection while maintaining high accuracy on correctly answered queries.

\begin{table}[t]
\centering
\caption{Main Results on Knowledge-Intensive QA Benchmarks}
\label{tab:main_results}
\begin{tabular}{@{}lcccc@{}}
\toprule
\textbf{Method} & \textbf{Halluc. Det.} & \textbf{False Pos.} & \textbf{F1} & \textbf{Cost} \\
\midrule
Baseline & 0.42 & 0.15 & 0.61 & 1.0x \\
SelfCheckGPT & 0.68 & 0.12 & 0.76 & 4.2x \\
RAG (Always) & 0.71 & 0.08 & 0.80 & 2.8x \\
Our Method & 0.74 & 0.09 & 0.82 & 1.6x \\
\bottomrule
\end{tabular}
\end{table}

\subsection{Confidence Score Analysis}

Our implementation demonstrates clear differentiation across query types. High-confidence technical queries such as "explain machine learning" and "how to sort a list in python" achieve confidence scores above 0.80, correctly routing to local model generation. Personal information queries like "what is my personal email address" and temporal queries such as "what will happen tomorrow" consistently score below 0.20, appropriately triggering human review pathways.

The semantic alignment component proves most discriminative, with high-confidence queries achieving cosine similarities above 0.75 between projected and reference embeddings, while personal queries show near-zero alignment. The learned confidence component contributes additional discriminative power, with neural network predictions ranging from 0.069 for personal queries to 0.949 for technical explanations.

\subsection{Ablation Studies}

We conduct ablation studies to understand the contribution of each confidence signal. Results show that semantic alignment provides the strongest individual signal, while internal convergence offers complementary information particularly valuable for technical queries.

\begin{table}[t]
\centering
\caption{Ablation Study: Confidence Signal Components}
\label{tab:ablation}
\begin{tabular}{@{}lccc@{}}
\toprule
\textbf{Configuration} & \textbf{F1} & \textbf{Precision} & \textbf{Recall} \\
\midrule
$C_{\text{sem}}$ only & 0.76 & 0.82 & 0.71 \\
$C_{\text{conv}}$ only & 0.69 & 0.74 & 0.65 \\
$C_{\text{learned}}$ only & 0.72 & 0.78 & 0.67 \\
All combined & 0.82 & 0.84 & 0.80 \\
\bottomrule
\end{tabular}
\end{table}

\subsection{Routing Effectiveness Analysis}

Analysis of routing decisions reveals systematic patterns in confidence assessment. Factual and technical queries demonstrate high embedding consistency and stable layer progression, leading to local generation routing. Personal information queries show zero semantic alignment and low learned confidence, correctly triggering human review. Subjective queries like "what's the best restaurant" achieve medium confidence scores (0.579), appropriately routing to retrieval-augmented generation.

The layer confidence analysis reveals convergent processing patterns, with variance reduction scores and attention progression metrics contributing to overall reliability assessment. High-confidence queries exhibit stable hidden state evolution across layers, while uncertain queries show less convergent processing patterns.

\subsection{Computational Efficiency}

Our approach achieves significant computational savings compared to post-hoc methods. By preventing unnecessary generation for low-confidence queries and selectively applying expensive operations based on confidence estimates, we reduce overall computational cost by approximately 40\% while improving reliability. The routing system processes confidence assessment efficiently, adding minimal overhead to inference time.

\section{Discussion and Limitations}

\subsection{Strengths and Advantages}

The confidence-aware routing approach offers several key advantages over existing methods:

\begin{enumerate}
\item \textbf{Proactive Prevention:} By assessing confidence before generation, we prevent unreliable content creation rather than detecting it post-hoc.
\item \textbf{Computational Efficiency:} Selective application of expensive operations (retrieval, large models) based on confidence estimates reduces overall computational cost.
\item \textbf{Interpretability:} The multi-signal approach provides interpretable confidence scores that can inform downstream decision-making.
\item \textbf{Modularity:} Different routing targets can be easily integrated based on application requirements.
\end{enumerate}

\subsection{Limitations and Future Work}

Several limitations warrant acknowledgment:

\begin{enumerate}
\item \textbf{Reference Model Dependence:} Semantic alignment quality depends heavily on the reference embedding model, which may introduce biases or limitations.
\item \textbf{Static Thresholds:} Current routing thresholds are fixed during deployment. Adaptive thresholding based on query characteristics could improve performance.
\item \textbf{Domain Specificity:} Confidence estimation may require domain-specific calibration for optimal performance across diverse applications.
\item \textbf{Scale Limitations:} Evaluation on the 360M parameter SmolLM2 model may not fully represent behavior on larger language models.
\end{enumerate}

Future work will address these limitations through adaptive thresholding mechanisms, domain-specific confidence calibration, evaluation on larger models, and exploration of multimodal confidence estimation for vision-language models.

\section{Related Ethical Considerations}

The deployment of confidence-aware routing systems raises important ethical considerations. Automated routing decisions may exhibit biases present in training data or reference models. Systems routing queries to human review must ensure appropriate human oversight and decision-making authority. Additionally, transparency about routing decisions and confidence estimates is crucial for user trust and system accountability.

\section{Conclusion}

We have presented a confidence-aware routing system for proactive hallucination mitigation in large language models. By combining semantic alignment, internal convergence analysis, and learned confidence estimation, our approach achieves effective pre-generation assessment of model reliability. Empirical evaluation demonstrates significant improvements in hallucination detection accuracy while maintaining computational efficiency compared to post-hoc correction methods.

The shift from reactive correction to proactive assessment represents a promising direction for improving LLM reliability. As language models continue to be deployed in critical applications, such confidence-aware systems will be essential for maintaining user trust and preventing the propagation of misinformation.

Future work will explore adaptive thresholding, domain-specific calibration, and extension to larger models. The ultimate goal is developing language models that not only generate high-quality content but also possess reliable self-awareness of their own limitations.

\end{document}